%% file: main.tex
\documentclass[10pt,twocolumn,letterpaper]{article}

\usepackage{cvpr}              
\usepackage{graphicx}
\usepackage{subcaption}

\input{preamble}

\definecolor{cvprblue}{rgb}{0.21,0.49,0.74}
\usepackage[pagebackref,breaklinks,colorlinks,citecolor=cvprblue]{hyperref}

\def\paperID{*****} 
\def\confName{CVPR}
\def\confYear{2024}

\title{Inpaint Biases:  A Pathway to Accurate and Unbiased Image Generation}

\author{Jiyoon Myung\\
MODULABS\\
{\tt\small jiyoon0424@gmail.com}
\and
Jihyeon Park\\
MODULABS\\
{\tt\small milhaud1201@gmail.com}
}

\begin{document}
\maketitle
\input{sec/0_abstract}    
\input{sec/1_intro}
\clearpage
\input{sec/3_preliminary_experiment}
\input{sec/4_method}

\input{sec/5_experiment}

\input{sec/6_conclusion}
\input{sec/7_future_research}

\section*{Acknowledgements}
This research was supported by Brian Impact Foundation, a non-profit organization dedicated to the advancement of science and technology for all.
{
    \small
    \bibliographystyle{ieeenat_fullname}
    \bibliography{main}
}

\end{document}


\section*{Supplementary Results}
\begin{longtable}{c|c|c}
\hline
\textbf{Initial Image (Initial Prompt)} & \textbf{Mask Image (Target Object)} & \textbf{Inpainted Image (Refined Prompt)} \\
\hline
\endfirsthead

\hline
\textbf{Initial Image (Initial Prompt)} & \textbf{Mask Image (Target Object)} & \textbf{Inpainted Image (Refined Prompt)} \\
\hline
\endhead

\hline
\endfoot

\hline
\endlastfoot

\begin{minipage}[t]{.25\textwidth}
    \vspace{3pt} 
  \centering
  \includegraphics[width=\textwidth]{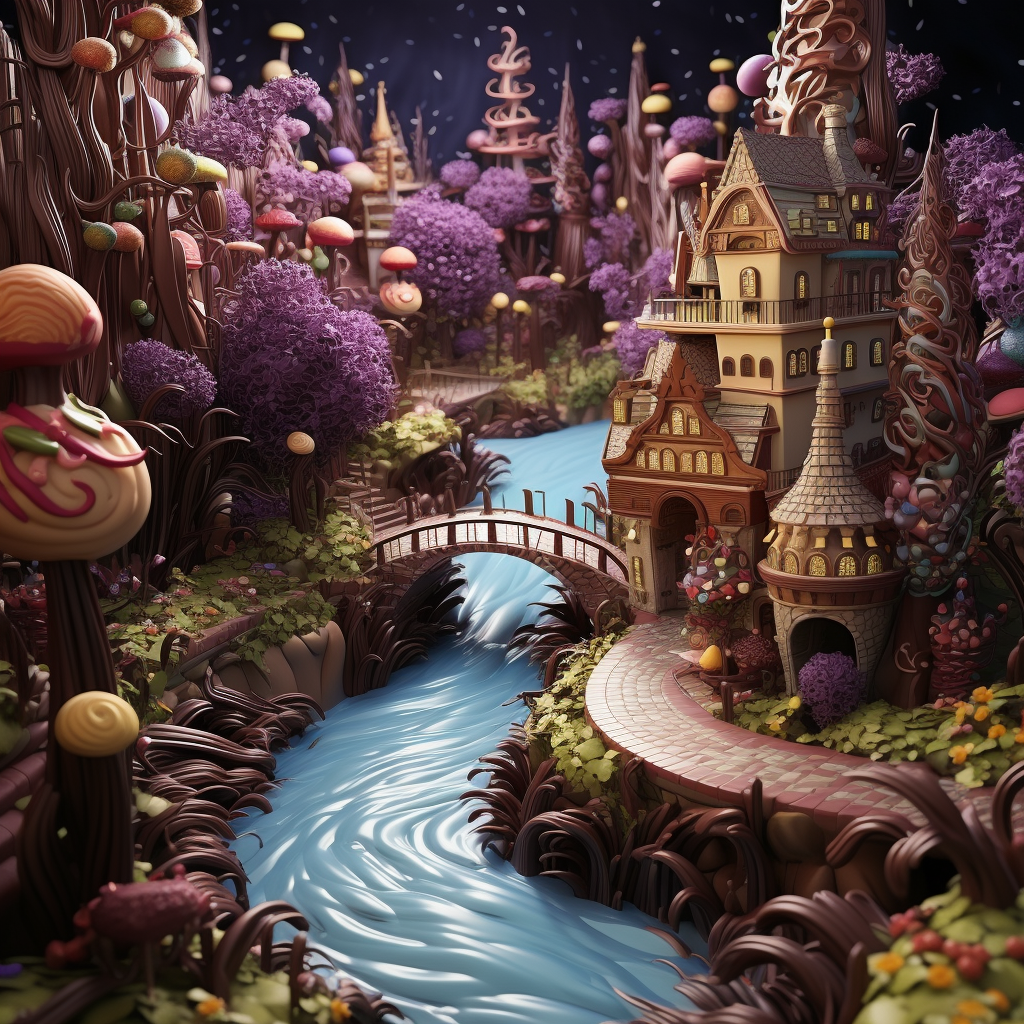}
  (a fantasy world where a river is made of dark chocolate)
  \vspace{5pt} 
\end{minipage} & 
\begin{minipage}[t]{.25\textwidth}
\vspace{3pt} 
  \centering
  \includegraphics[width=\textwidth]{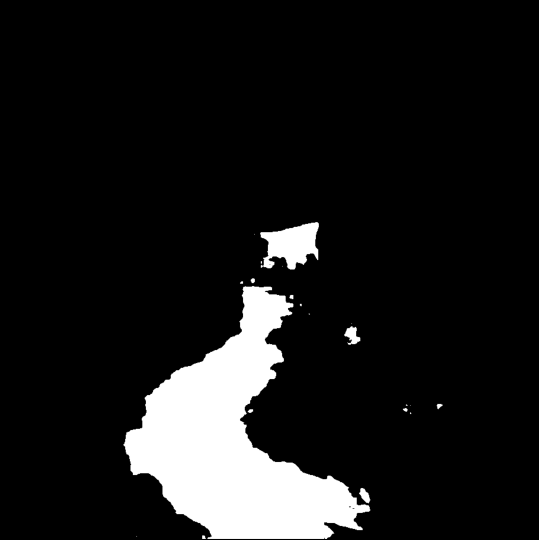}
  (a standard blue river)
  \vspace{5pt} 
\end{minipage} & 
\begin{minipage}[t]{.25\textwidth}
\vspace{3pt} 
  \centering
  \includegraphics[width=\textwidth]{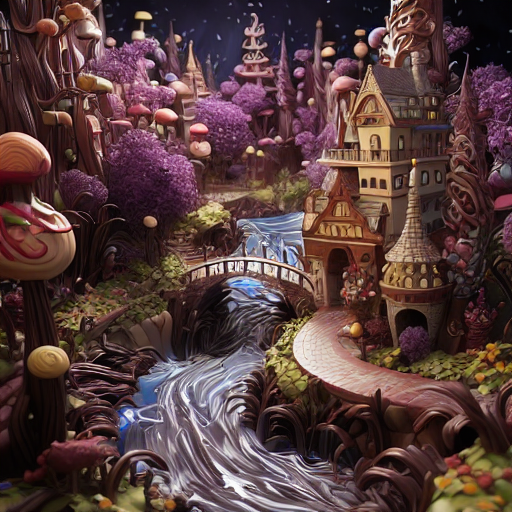}
  (a smooth, flowing dark chocolate texture, rich and velvety, blending seamlessly into a whimsical, candy-themed fantasy landscape)
  \vspace{5pt} 
\end{minipage} \\
\hline

\begin{minipage}[t]{.25\textwidth}
    \vspace{3pt} 
  \centering
  \includegraphics[width=\textwidth]{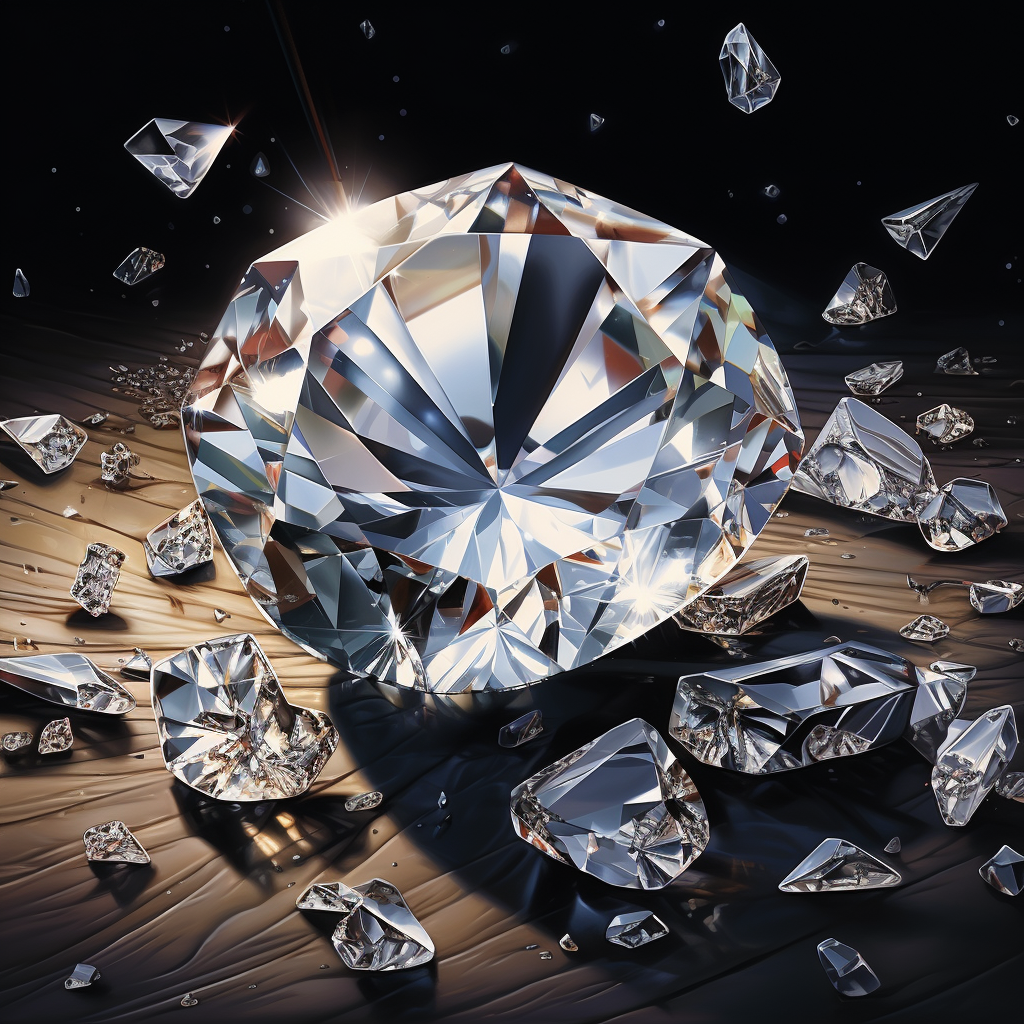}
  (diamonds broke into pieces)
  \vspace{5pt} 
\end{minipage} & 
\begin{minipage}[t]{.25\textwidth}
\vspace{3pt} 
  \centering
  \includegraphics[width=\textwidth]{assets/diamond_mask.png}
  (a large, intact diamond)
  \vspace{5pt} 
\end{minipage} & 
\begin{minipage}[t]{.25\textwidth}
\vspace{3pt} 
  \centering
  \includegraphics[width=\textwidth]{assets/diamond_inpainted.png}
  (small shredded pieces of diamond)
  \vspace{5pt} 
\end{minipage} \\
\hline

\begin{minipage}[t]{.25\textwidth}
\vspace{3pt} 
  \centering
  \includegraphics[width=\textwidth]{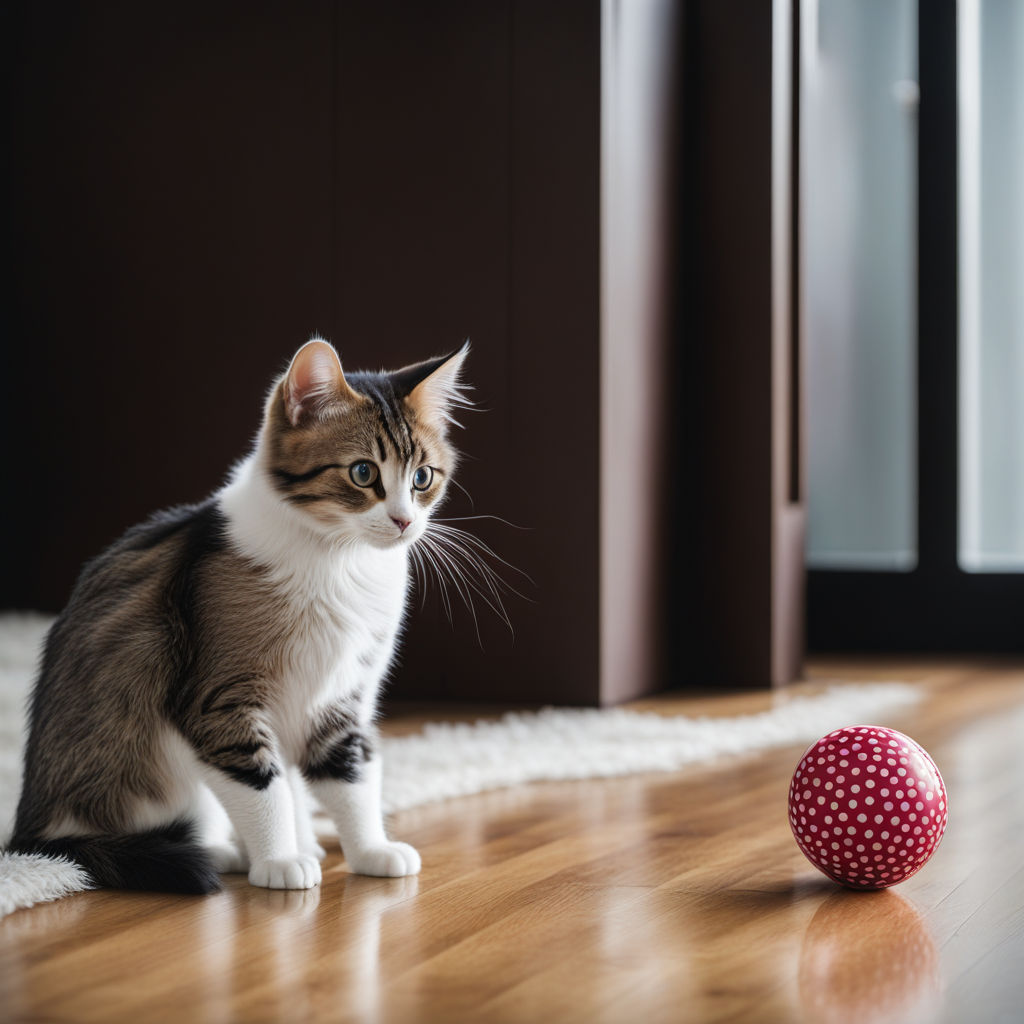}
  (a cat with a polka-dotted fur pattern playing a toy)
  \vspace{5pt}
\end{minipage} & 
\begin{minipage}[t]{.25\textwidth}
\vspace{3pt} 
  \centering
  \includegraphics[width=\textwidth]{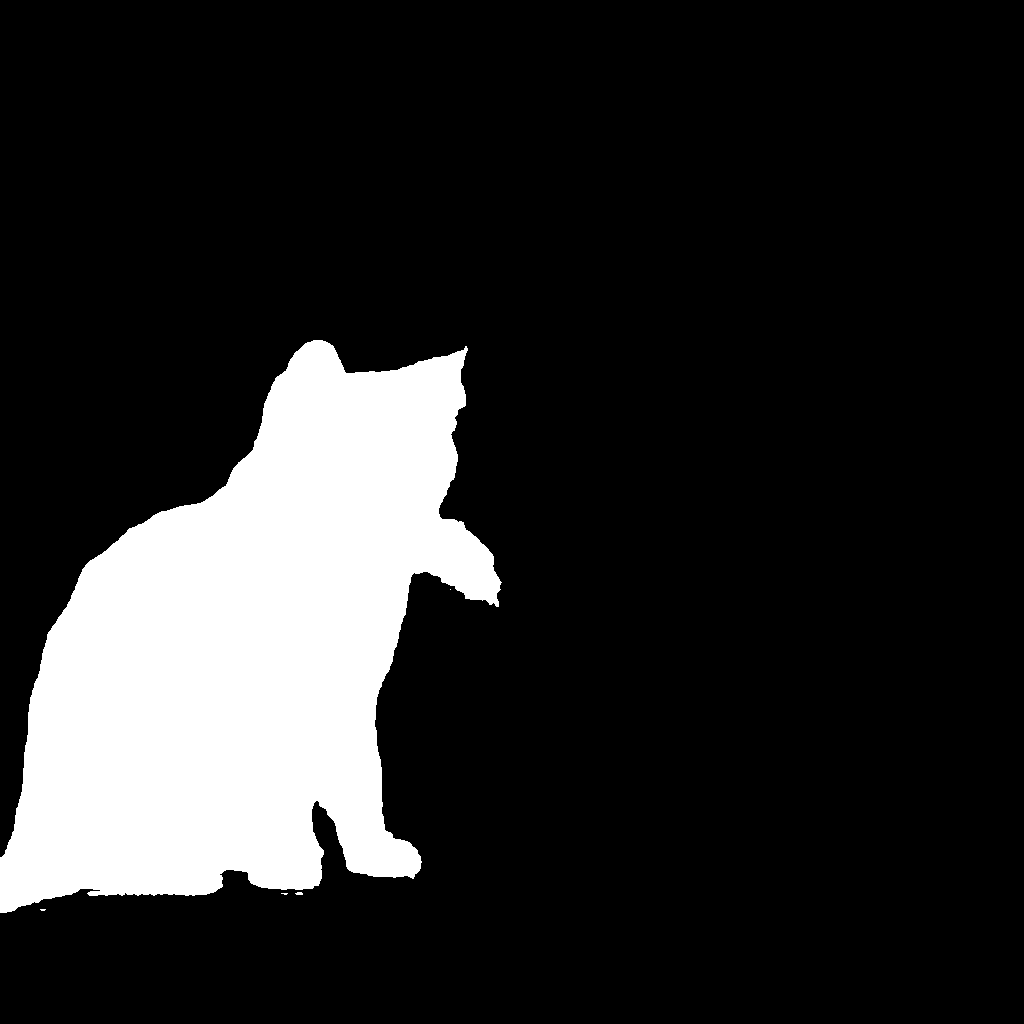}
  (a conventional cat without pattern)
  \vspace{5pt}
\end{minipage} & 
\begin{minipage}[t]{.25\textwidth}
\vspace{3pt} 
  \centering
  \includegraphics[width=\textwidth]{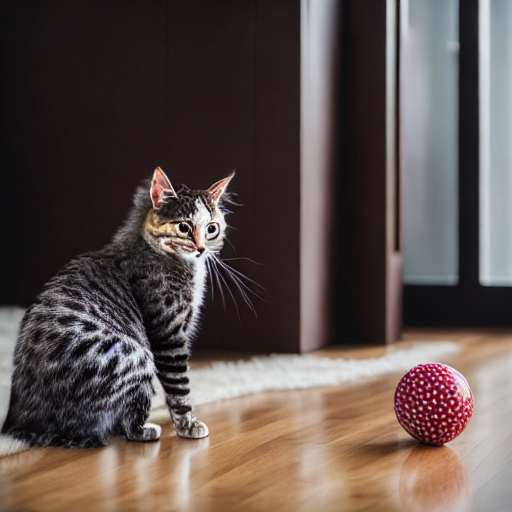}
  (an imaginative depiction of a polka-dotted cat highlighting its unique fur pattern)
  \vspace{5pt}
\end{minipage} \\
\hline

\begin{minipage}[t]{.25\textwidth}
\vspace{3pt} 
  \centering
  \includegraphics[width=\textwidth]{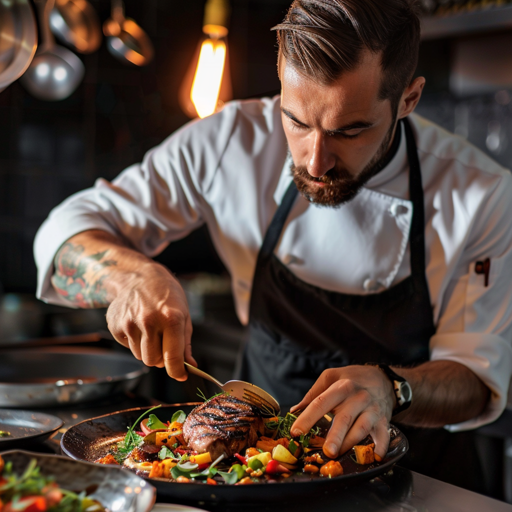}
  (a chef prepares a delicious meal)
  \vspace{5pt}
\end{minipage} & 
\begin{minipage}[t]{.25\textwidth}
\vspace{3pt} 
  \centering
  \includegraphics[width=\textwidth]{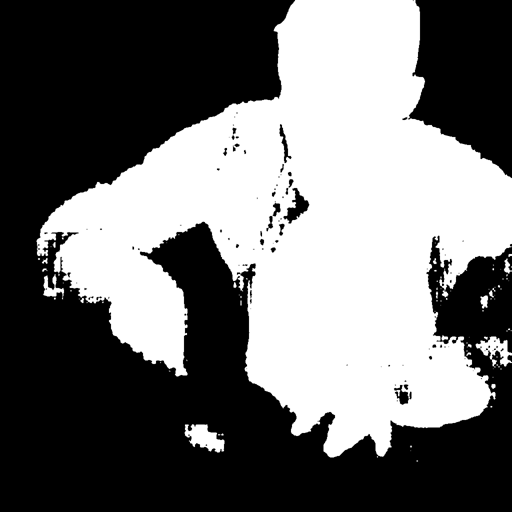}
  (a Western chef)
  \vspace{5pt}
\end{minipage} & 
\begin{minipage}[t]{.25\textwidth}
\vspace{3pt} 
  \centering
  \includegraphics[width=\textwidth]{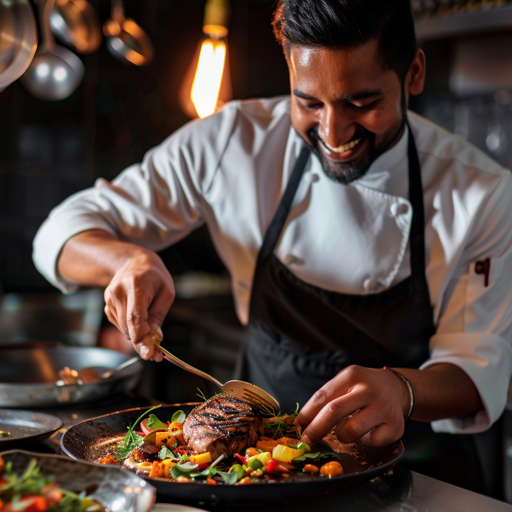}
  (a South Asian male chef with a warm smile prepares a meal)
  \vspace{5pt}
\end{minipage} \\
\hline

\begin{minipage}[t]{.25\textwidth}
\vspace{3pt} 
  \centering
  \includegraphics[width=\textwidth]{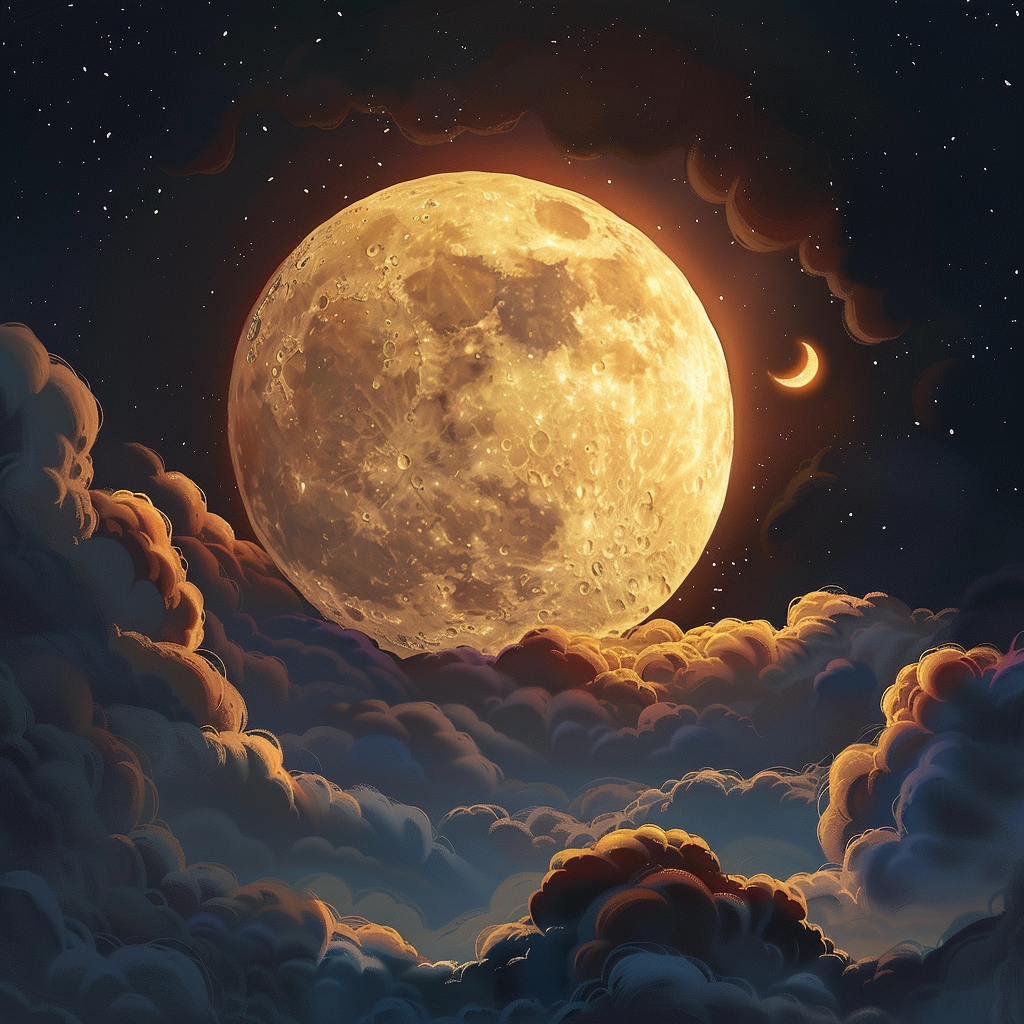}
  (chocolate chip cookie moon with clouds)
  \vspace{5pt}
\end{minipage} & 
\begin{minipage}[t]{.25\textwidth}
\vspace{3pt} 
  \centering
  \includegraphics[width=\textwidth]{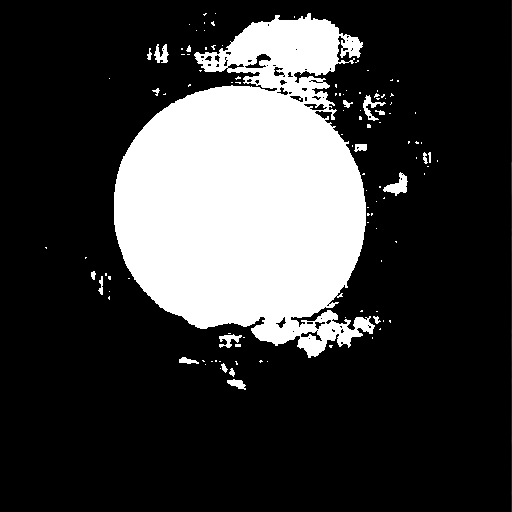}
  (a conventional moon)
  \vspace{5pt}
\end{minipage} & 
\begin{minipage}[t]{.25\textwidth}
\vspace{3pt} 
  \centering
  \includegraphics[width=\textwidth]{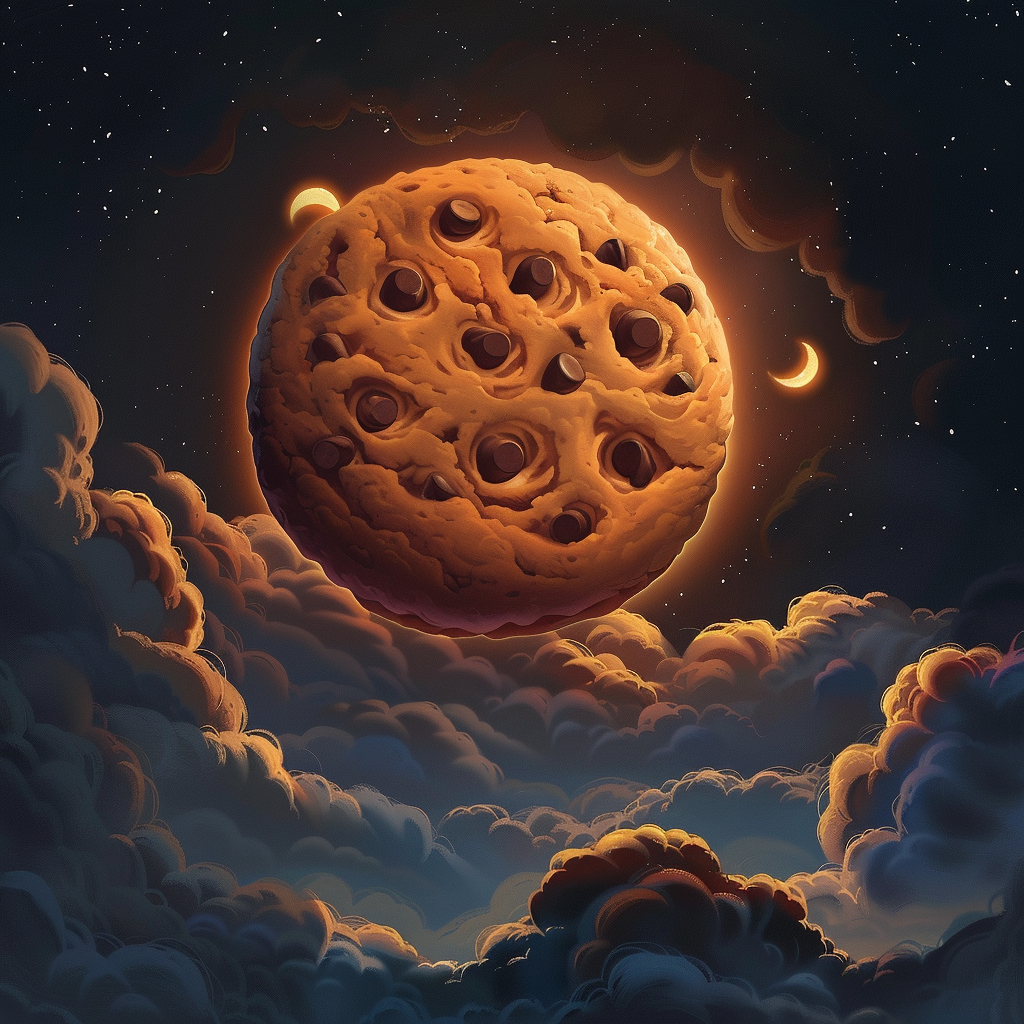}
  (warm, golden-brown, soft center, melted chocolate chips, homemade, fresh from oven, slightly crispy edges, comforting, classic American dessert, inviting aroma)
  \vspace{5pt}
\end{minipage} \\
\hline

\begin{minipage}[t]{.25\textwidth}
\vspace{3pt} 
  \centering
  \includegraphics[width=\textwidth]{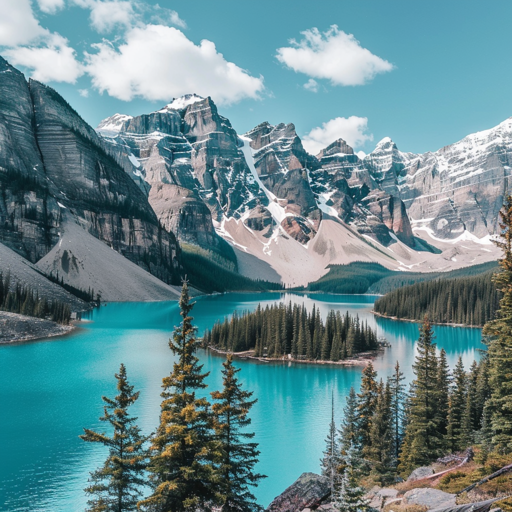}
  (a photo of rocky mountains like ice cream)
  \vspace{5pt}
\end{minipage} & 
\begin{minipage}[t]{.25\textwidth}
\vspace{3pt} 
  \centering
  \includegraphics[width=\textwidth]{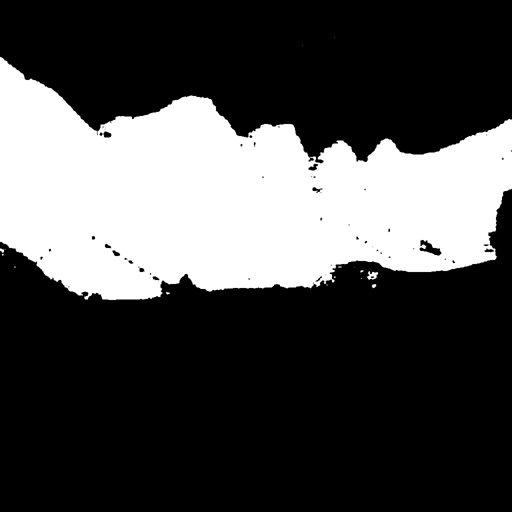}
  (conventional rocky mountains)
  \vspace{5pt}
\end{minipage} & 
\begin{minipage}[t]{.25\textwidth}
\vspace{3pt} 
  \centering
  \includegraphics[width=\textwidth]{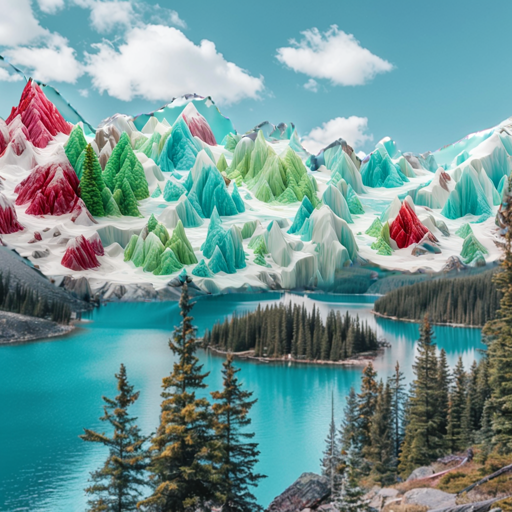}
  (majestic peaks, creamy textures, vibrant colors, chocolate swirls, vanilla snowcaps, mint green valleys, cherry toppings, confectionery fantasy)
  \vspace{5pt}
\end{minipage} \\
\hline

\end{longtable}

%% file: preamble.tex
\usepackage[dvipsnames]{xcolor}


%% file: sec/0_abstract.tex
\begin{abstract}
This paper examines the limitations of advanced text-to-image models in accurately rendering unconventional concepts which are scarcely represented or absent in their training datasets. We identify how these limitations not only confine the creative potential of these models but also pose risks of reinforcing stereotypes. To address these challenges, we introduce the Inpaint Biases framework, which employs user-defined masks and inpainting techniques to enhance the accuracy of image generation, particularly for novel or inaccurately rendered objects. Through experimental validation, we demonstrate how this framework significantly improves the fidelity of generated images to the user's intent, thereby expanding the models' creative capabilities and mitigating the risk of perpetuating biases. Our study contributes to the advancement of text-to-image models as unbiased, versatile tools for creative expression.
\end{abstract}

%% file: sec/1_intro.tex
\section{Introduction}
\label{sec:intro}

The advent of sophisticated text-to-image models \cite{rombach2022highresolution, ramesh2021zeroshot, saharia2022photorealistic}  has substantially transformed the landscape of creative content generation, becoming a crucial component in this domain. As these models become increasingly influential, it is crucial to evaluate the content they produce because any inherent biases or defects could have widespread implications. 

This paper delves into a critical examination of text-to-image models' difficulties in rendering unconventional concepts, like blue bananas or square watermelons, which are barely represented or absent in their training data. This limitation not only constricts the creative potential of the models but also poses the risk of leading to stereotypical representations of various demographics or societal roles. For example, consistently misrepresenting or failing to acknowledge specific cultural or societal elements could reinforce or exacerbate existing social and cultural inequalities. Moreover, using images generated by these models as training data could further solidify these biases, creating a feedback loop of biased content generation.

Addressing this, our study aims to develop methods that enable text-to-image models to more accurately depict unusual and innovative concepts. Our exploration involved experimenting with a range of prompts to assess how these models handle uncommon or novel concepts. Although the models efficiently generated these unique concepts in isolation, they struggled to maintain accuracy when these concepts were integrated with standard objects or within complex style prompts. This observation prompted us to devise a method that specifically enhances the portrayal of incorrectly rendered, novel objects through focused inpainting techniques. 

To deal with this concern, we introduce the Inpaint Biases framework that leverages user-created masks and inpainting techniques for precise corrections within images. This process initiates with generating an image from a user's prompt. If the image inaccurately reflects the intended concept, such as rendering a blue banana as yellow, the user can apply a mask to the incorrect element, directing the model's correction efforts specifically to this area. Enhanced prompts, refined using large language models (LLM), are then focused on the masked region to improve accuracy and detail, followed by an inpainting process that melds the corrections with the original image seamlessly.

By refining and inpainting these areas, the Inpaint Biases framework significantly narrows the gap between the generated image and the user's original intention, maintaining the overall aesthetic while meticulously correcting the inaccuracies. This approach not only expands the creative capabilities of text-to-image models but also plays a pivotal role in mitigating the risk of perpetuating biases, steering the technology towards a future where it can serve as an unbiased, versatile tool for artistic and creative expression. Through this framework, we are taking a significant step toward harnessing the full potential of text-to-image models in a manner that is both creatively enriching and socially responsible.

%% file: sec/3_preliminary_experiment.tex
\section{Exploring Bias Towards Unusual Concepts}
\begin{figure}[]
    \centering
    \begin{subfigure}[b]{0.23\columnwidth}
        \centering
        \includegraphics[width=\textwidth]{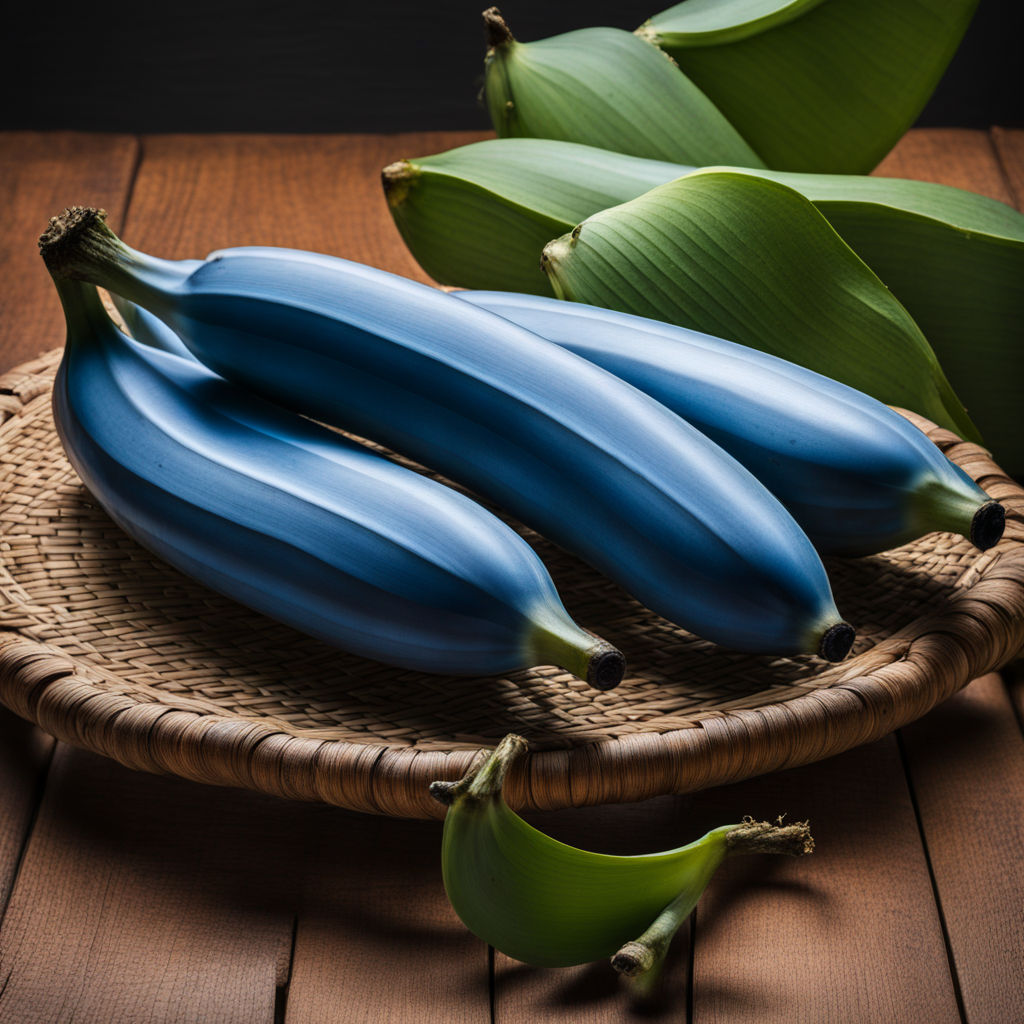}
        \label{fig:first}
    \end{subfigure}
    \begin{subfigure}[b]{0.23\columnwidth}
        \centering
        \includegraphics[width=\textwidth]{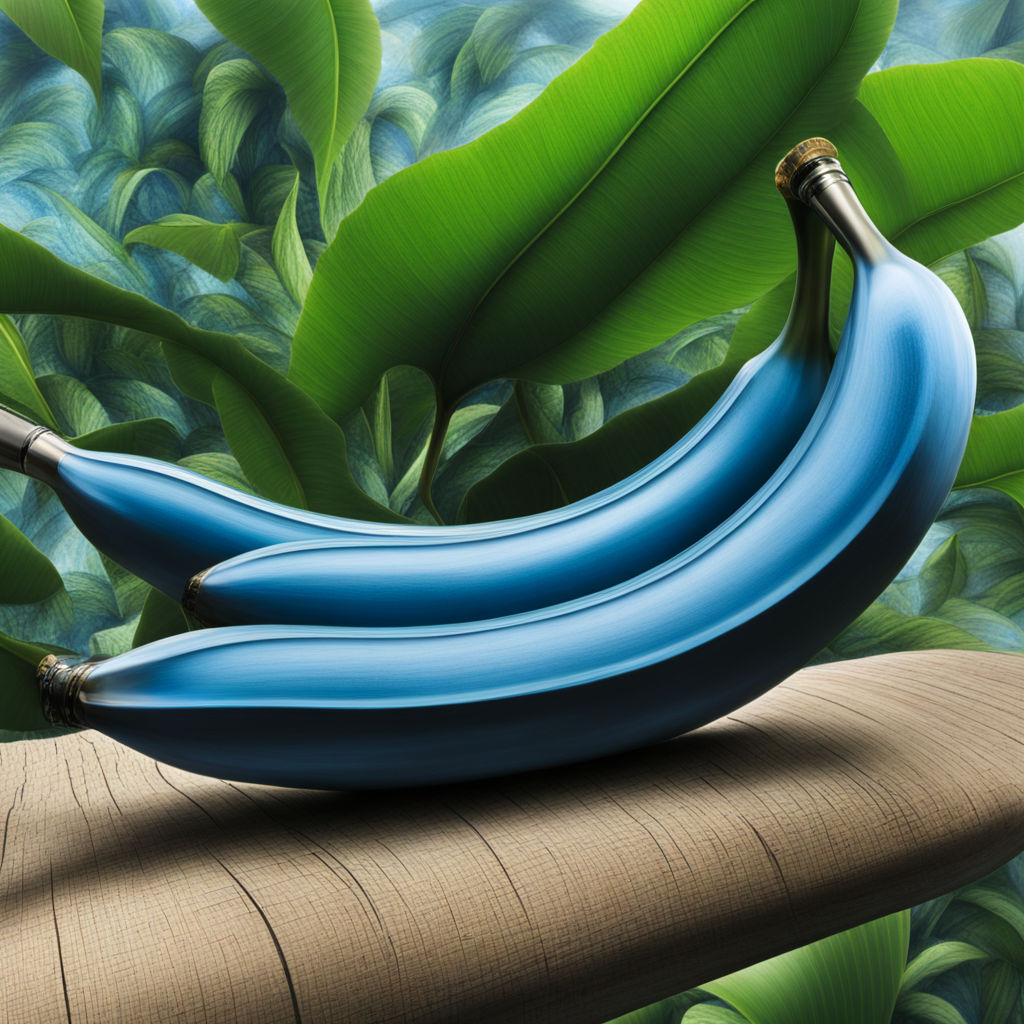}
        \label{fig:second}
    \end{subfigure}
    \begin{subfigure}[b]{0.23\columnwidth}
        \centering
        \includegraphics[width=\textwidth]{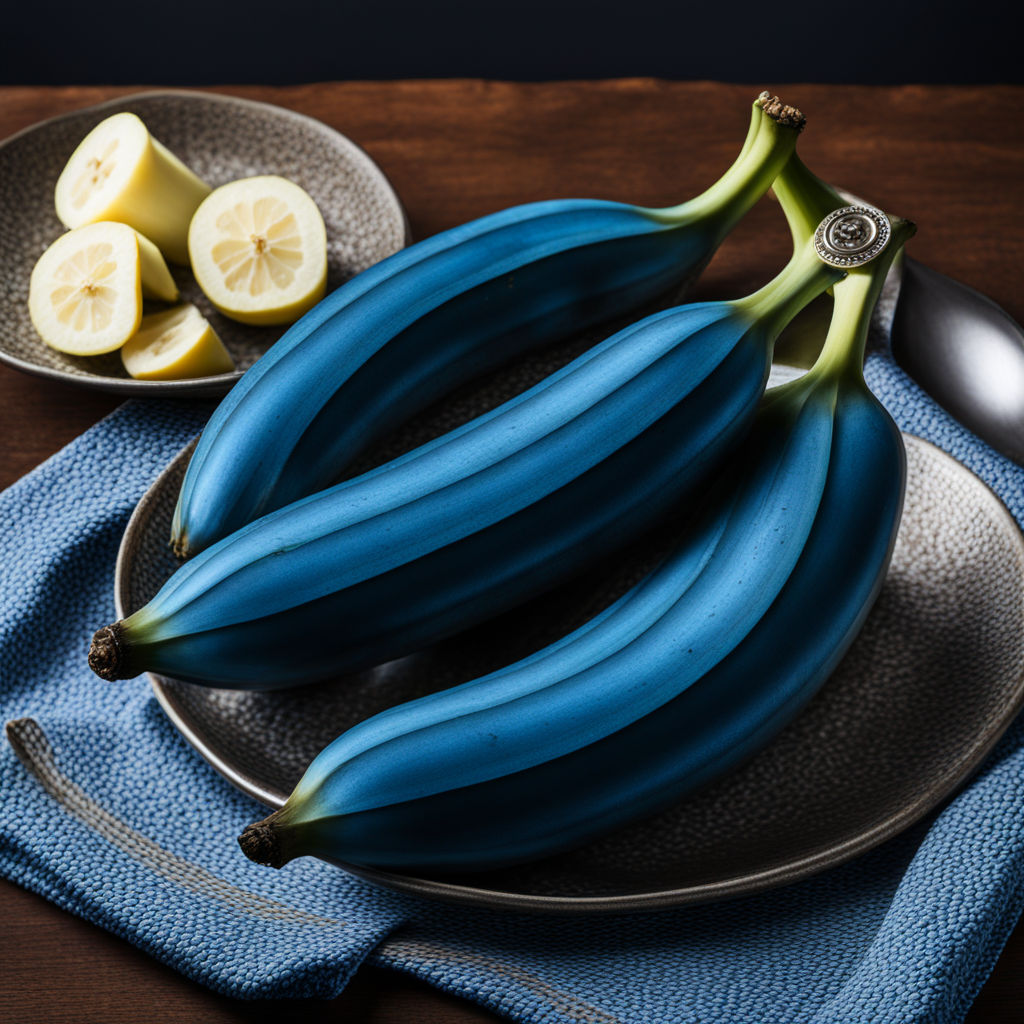}
        \label{fig:third}
    \end{subfigure}
    \begin{subfigure}[b]{0.23\columnwidth}
        \centering
        \includegraphics[width=\textwidth]{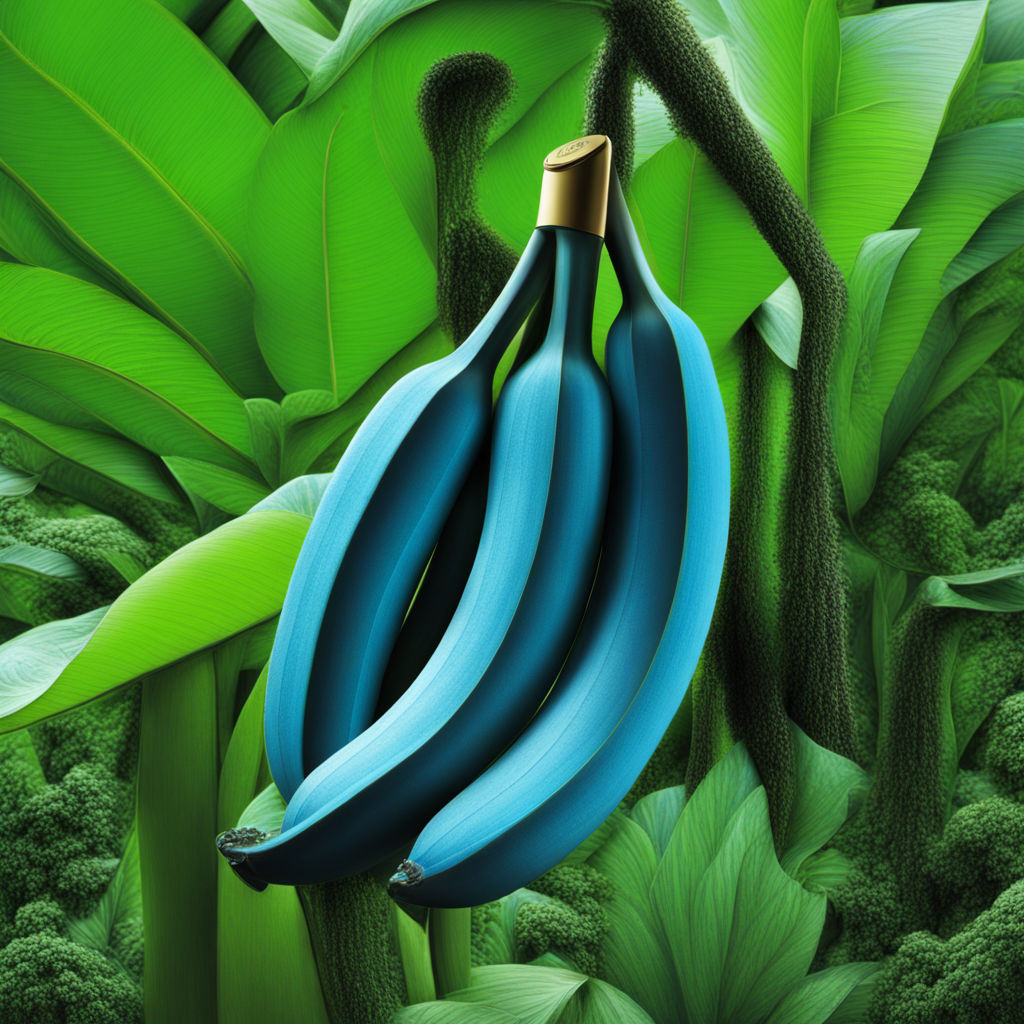}
        \label{fig:fourth}
    \end{subfigure}

    \caption{Illustrations of 'blue bananas'.}
    \label{fig:isolation}
\end{figure}

\begin{figure}[]
    \centering
    \begin{subfigure}[b]{0.23\columnwidth}
        \centering
        \includegraphics[width=\textwidth]{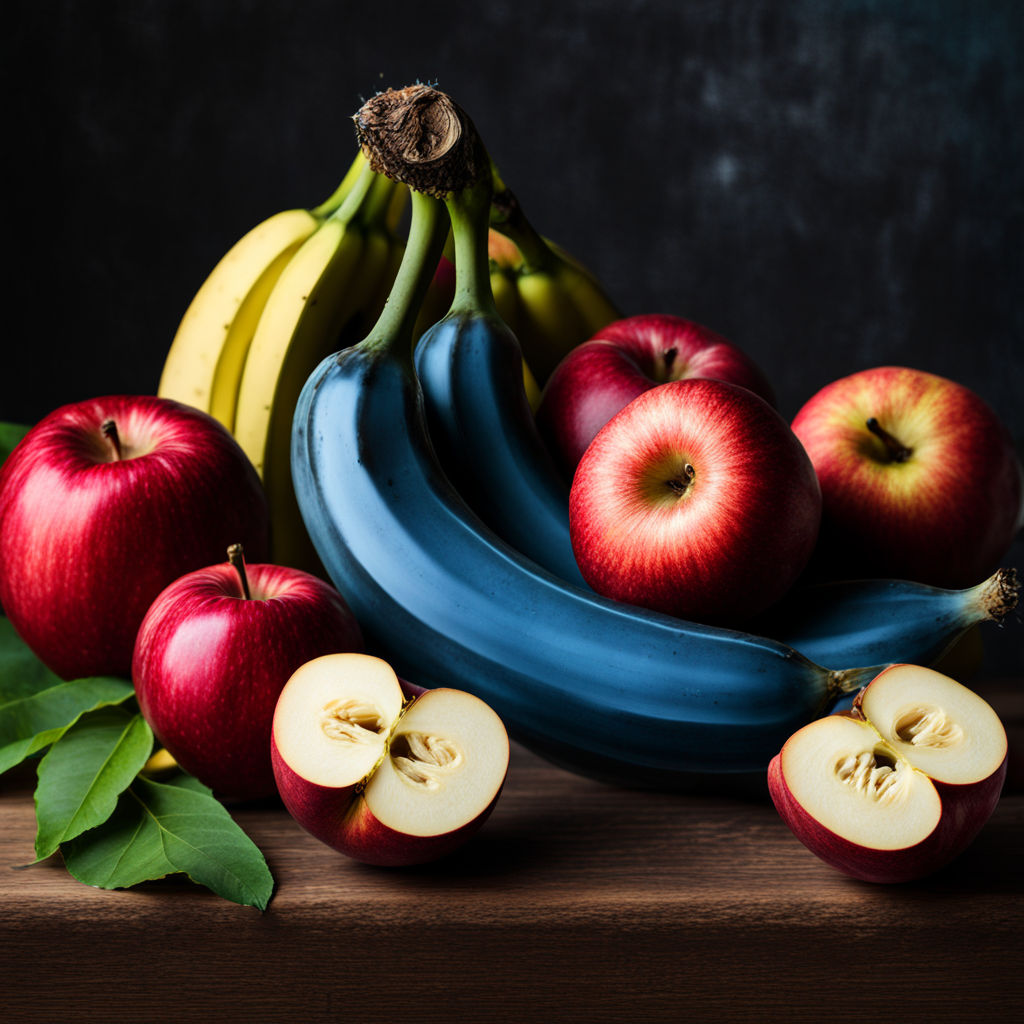}
        \label{fig:first}
    \end{subfigure}
    \begin{subfigure}[b]{0.23\columnwidth}
        \centering
        \includegraphics[width=\textwidth]{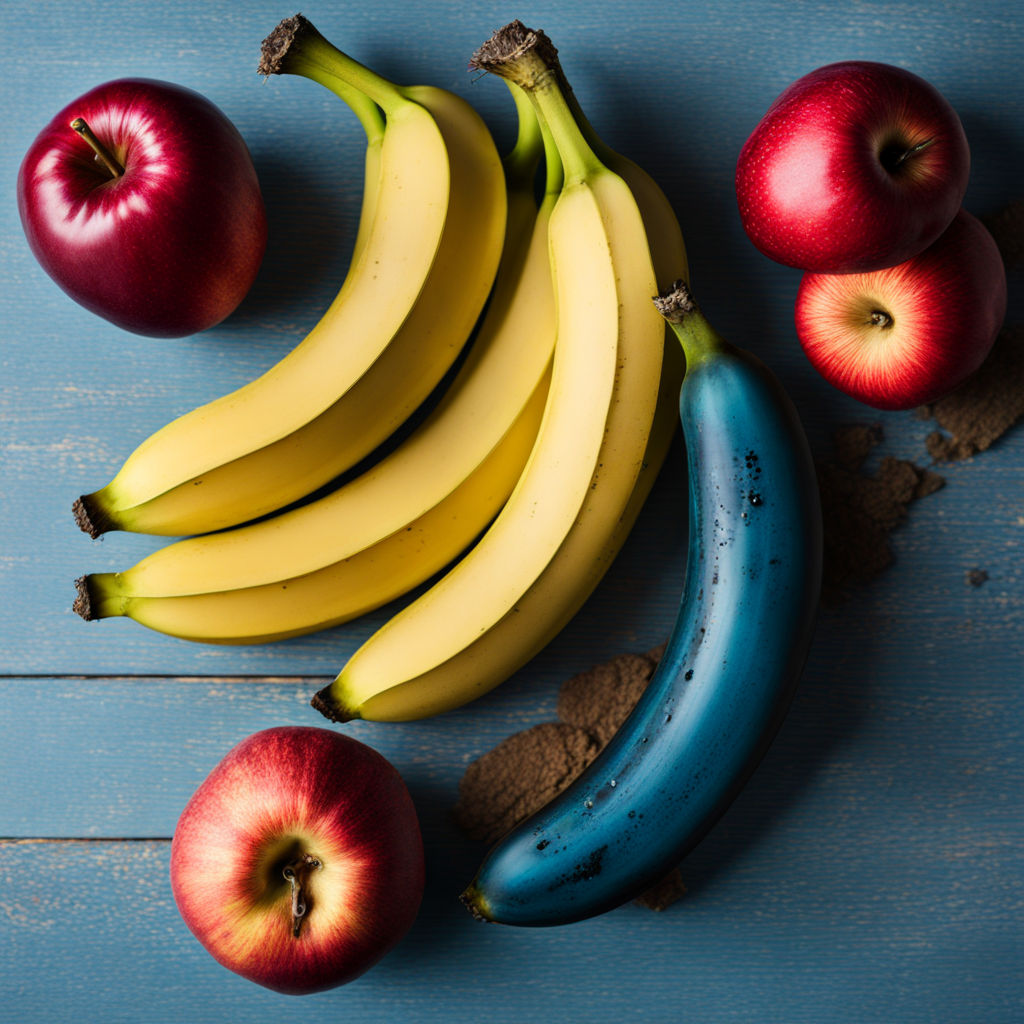}
        \label{fig:second}
    \end{subfigure}
    \begin{subfigure}[b]{0.23\columnwidth}
        \centering
        \includegraphics[width=\textwidth]{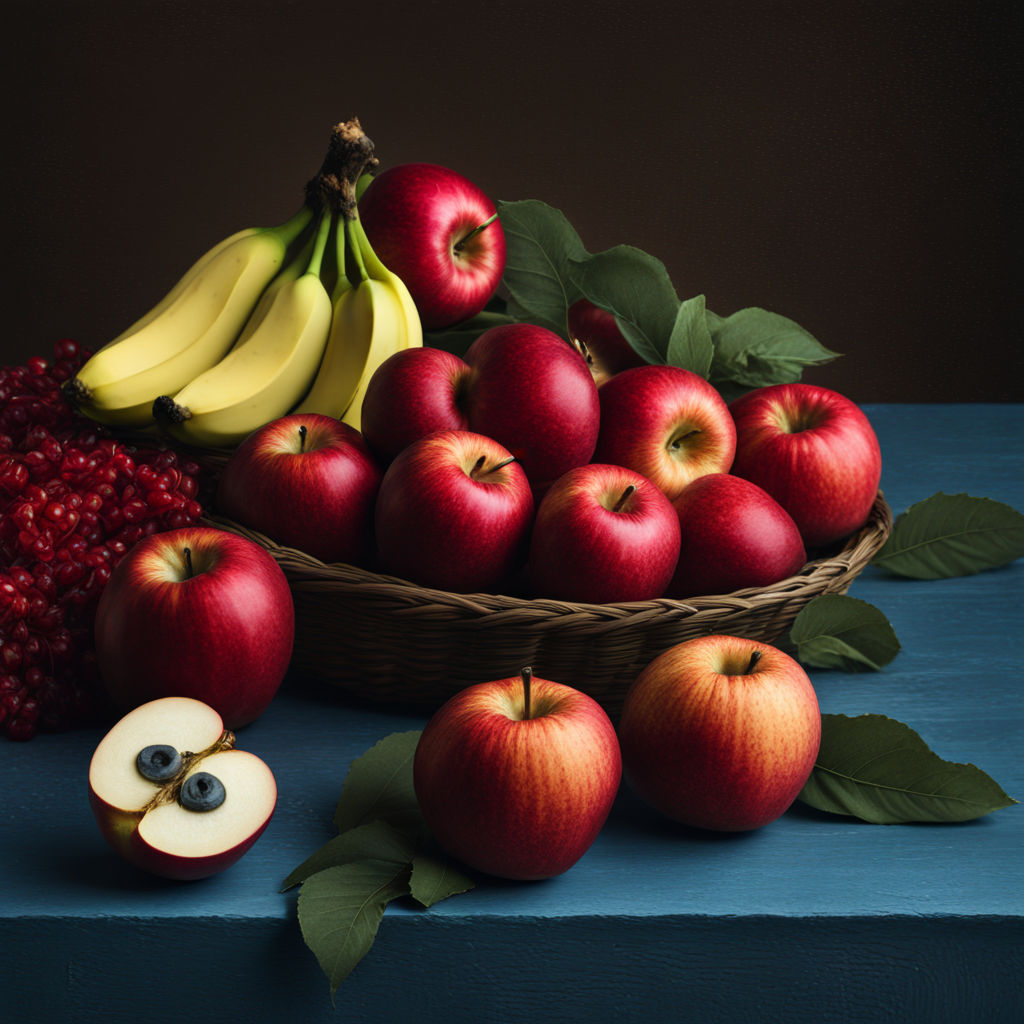}
        \label{fig:third}
    \end{subfigure}
    \begin{subfigure}[b]{0.23\columnwidth}
        \centering
        \includegraphics[width=\textwidth]{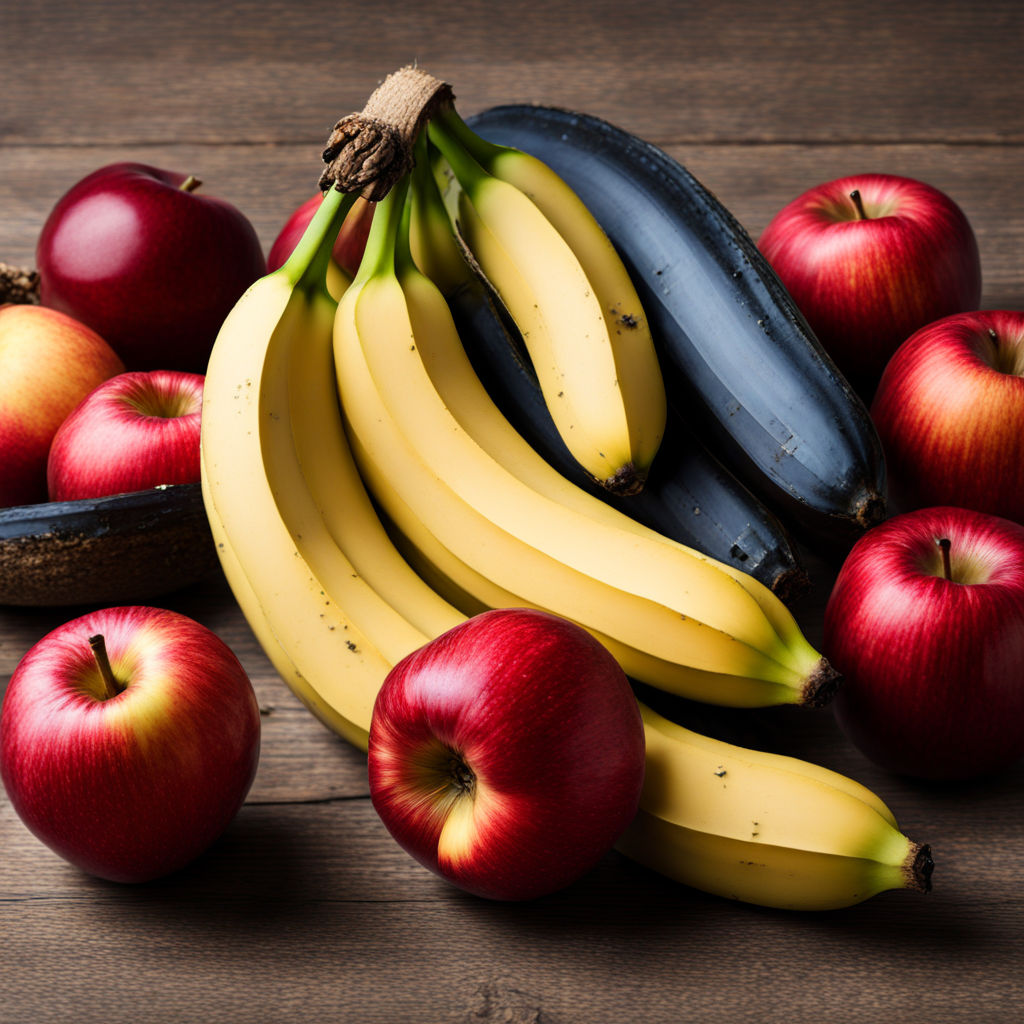}
        \label{fig:fourth}
    \end{subfigure}

    \caption{Illustrations of 'blue bananas and red apples on the table'.}
    \label{fig:mix}
\end{figure}

In our experimentation with text-to-image models, we systematically tested various prompts to assess the models’ ability to generate images based on uncommon concepts, such as 'blue bananas'. The results indicated that when tasked with rendering these unique concepts in isolation, the tool generally produced accurate and satisfactory images (Figure. \ref{fig:isolation}). However, when we introduced a combination of concepts, such as 'blue bananas and red apples on the table,' we observed a noticeable decline in the models’ ability to accurately depict the less common concept (the blue banana) alongside more conventional objects (Figure. \ref{fig:mix}).

This phenomenon suggests that when multiple concepts are included in a prompt, the tool may prioritize accurately rendering more common or familiar objects, consequently compromising the depiction of rarer concepts. Based on these findings, we deduced that a targeted approach focusing on the specific refinement of inaccurately rendered elements could enhance the overall image quality. Consequently, we propose an inpainting method that specifically addresses and corrects the portions of the image that are not accurately rendered.

By applying this inpainting technique, we could selectively refine the depiction of the 'blue bananas' without needing to regenerate the entire image. This approach not only improved the accuracy of the uncommon concept's representation but also preserved the integrity of the overall image. Therefore, we conclude that an inpainting strategy, which allows for the focused correction of inaccurately depicted elements, offers an effective solution for enhancing the text-to-image models’ performance, particularly when dealing with a mix of common and uncommon concepts within a single prompt.

%% file: sec/4_method.tex
\section{Inpaint Biases Framework}

\begin{figure*}[]
  \centering
  \includegraphics[width=\textwidth]{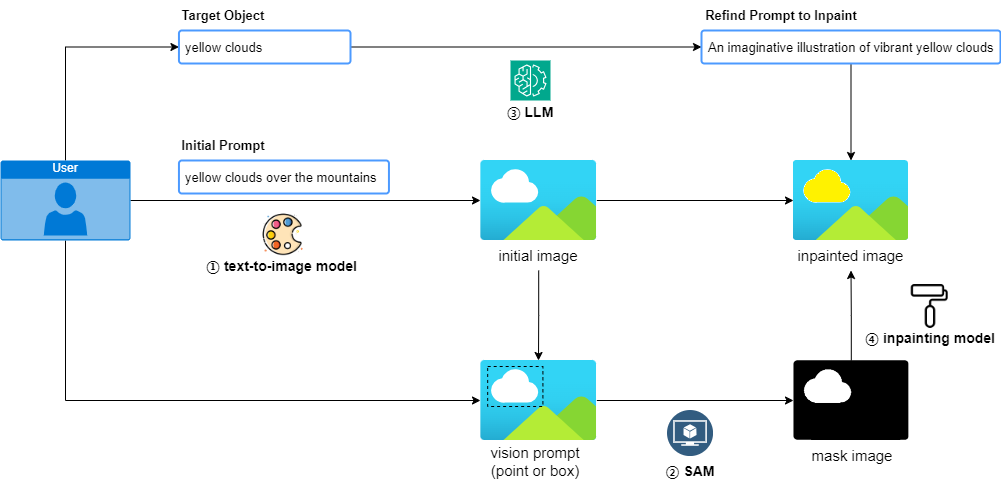}
   \caption{Inpaint Biases Framework}
   \label{method}
\end{figure*}

The Inpaint Biases framework is strategically developed to improve image generation by inpainting targeted areas. The framework marks an important first step toward addressing biases in image generation, aiming to minimize the discrepancy between the user's initial intent and the resultant image.

\subsection{Image Generation from Initial Prompt}

The initial step in the Inpaint Biases framework involves generating an image based on a user-defined initial prompt using a text-to-image tool. This prompt is expected to encompass an atypical concept, such as ‘yellow clouds over the mountains’, which standard image generation processes might not accurately capture.

\subsection{Mask Generation Based on User Input}

In instances where the generated image inadequately represents the unusual concept (yellow clouds), the user intervenes to delineate the target area for modification. This is achieved by selecting a specific point or defining a bounding box on the image. Subsequently, utilizing the zero-shot segmentation capabilities of the Segment Anything Model (SAM) \cite{kirillov2023segment}, a mask of the targeted area is created, setting the stage for focused inpainting. To enhance this process, users have the additional capability of selecting the target area by painting it themselves. This intuitive interaction allows for more precise and personalized input, ensuring that the modification aligns closely with the user's vision. 

\subsection{Prompt Refinement for Targeted Inpainting}

The third step diverges from the broader image generation to concentrate on the targeted concept. Here, the Large Language Model (LLM) plays a crucial role by suggesting a refined prompt that encapsulates the essence of the concept needing precise representation. For instance, if the user's objective is to depict 'yellow clouds' accurately, the LLM might suggest a detailed prompt like 'An imaginative illustration of vibrant yellow clouds,' facilitating a more targeted and conceptually aligned inpainting process.

\subsection{Inpainting Using the Refined Prompt}

With the newly refined prompt provided by the LLM, the final step is to inpaint the previously masked area, thereby aligning the unusual concept with the user's vision. This inpainting is executed with the intent to accurately fill in the masked section, reflecting the detailed attributes outlined in the refined prompt. This precision is achieved using a mask-generation strategy presented by LAMA \cite{suvorov2021resolutionrobust}, where the latent VAE (Variational Autoencoder) representations of the masked image are utilized for additional conditioning.

%% file: sec/5_experiment.tex
\section{Methodology Implementation}
In this section, we illustrate the application of the Inpaint Biases framework through a series of implementation examples, showcasing how this methodology can effectively correct inaccuracies in generated images. We selected a range of concepts that typically challenge text-to-image models to test the framework's capability to refine and correct these complex visual concepts.

\begin{figure}[!h]
\centering
\begin{subfigure}[b]{0.32\columnwidth}
    \includegraphics[width=\textwidth]{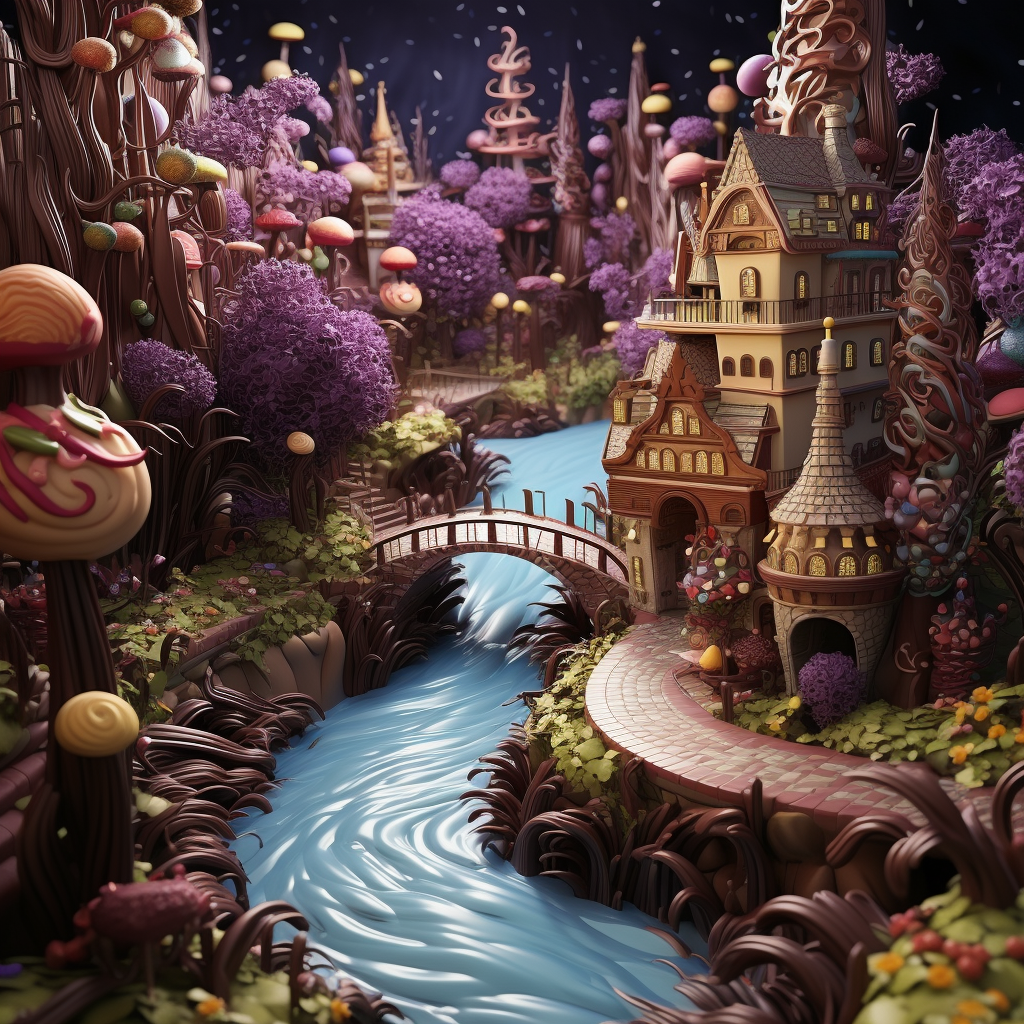}
    \caption{Initial Image}
    \label{fig:chocolate}
\end{subfigure}
\hfill
\begin{subfigure}[b]{0.32\columnwidth}
    \includegraphics[width=\textwidth]{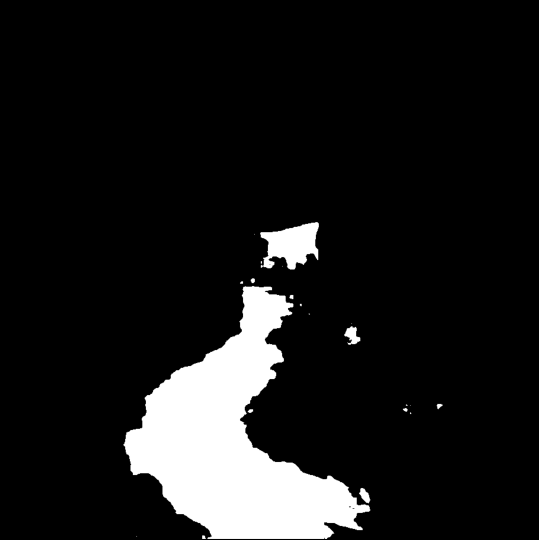}
    \caption{Mask Image}
    \label{fig:chocolate_mask}
\end{subfigure}
\hfill
\begin{subfigure}[b]{0.32\columnwidth}
    \includegraphics[width=\textwidth]{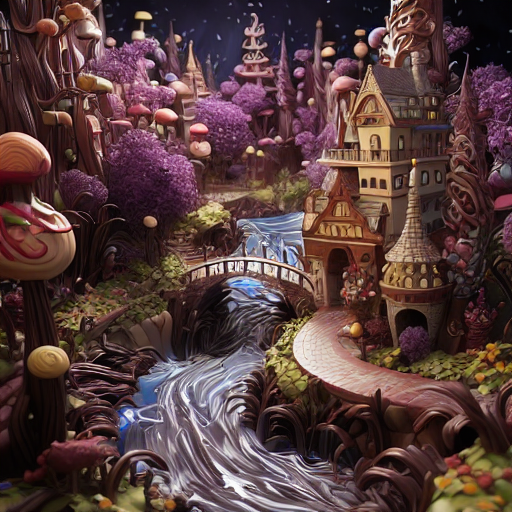}
    \caption{Inpainted Image}
    \label{fig:chocolate_inpainted}
\end{subfigure}
\caption{Example 1: chocolate river}
\label{sample1}
\end{figure}

As an example, we provide a compelling example with Figure. \ref{sample1}, where the initial image (Figure. \ref{fig:chocolate}) was generated from the prompt 'A fantasy world where a river is made of dark chocolate.' Despite the creative intent, the text-to-image model defaulted to a blue river, influenced by its training on prevalent images of rivers in standard blue. To address this discrepancy, we utilized a mask (Figure. \ref{fig:chocolate_mask}) to isolate the river area and then applied our inpainting process with a refined prompt: 'a smooth, flowing dark chocolate texture, rich and velvety, blending seamlessly into a whimsical, candy-themed fantasy landscape,' as suggested by the LLM. The result was a striking transformation, with the river convincingly altered to resemble flowing dark chocolate, aligning with the original creative vision (Figure. \ref{fig:chocolate_inpainted}). 

\begin{figure}[!h]
\centering
\begin{subfigure}[b]{0.32\columnwidth}
    \includegraphics[width=\textwidth]{assets/diamond.png}
    \caption{Initial Image}
    \label{fig:diamond}
\end{subfigure}
\hfill
\begin{subfigure}[b]{0.32\columnwidth}
    \includegraphics[width=\textwidth]{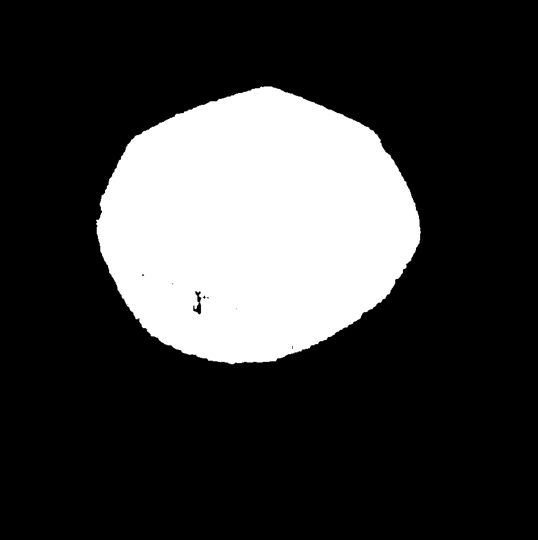}
    \caption{Mask Image}
    \label{fig:diamond_mask}
\end{subfigure}
\hfill
\begin{subfigure}[b]{0.32\columnwidth}
    \includegraphics[width=\textwidth]{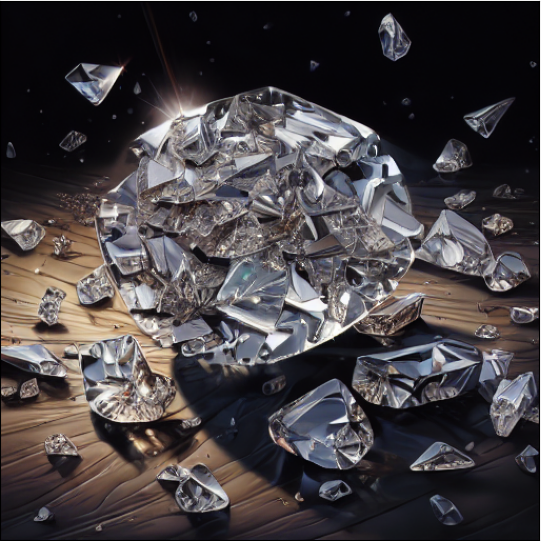}
    \caption{Inpainted Image}
    \label{fig:diamond_inpainted}
\end{subfigure}
\caption{Example 2: broken diamonds}
\label{sample2}
\end{figure}

\begin{figure}[!h]
\centering
\begin{subfigure}[b]{0.32\columnwidth}
    \includegraphics[width=\textwidth]{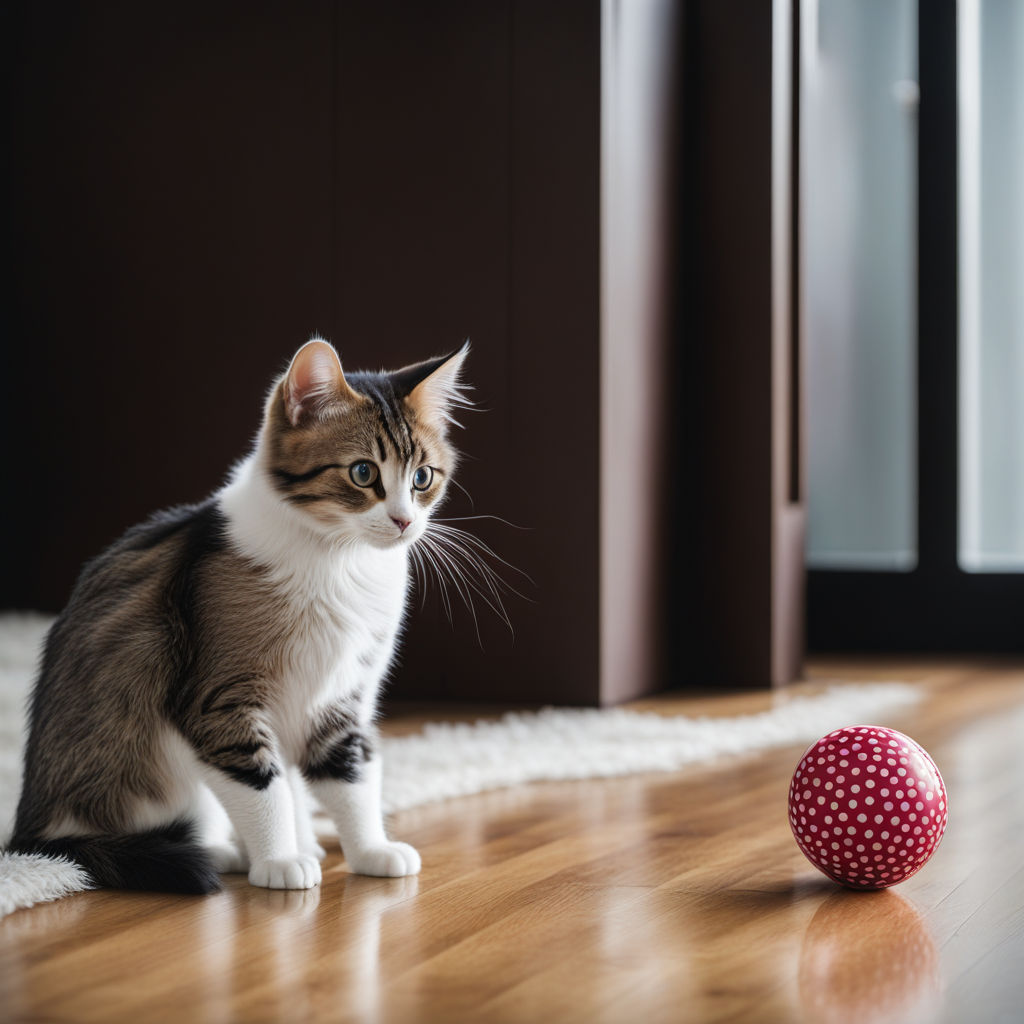}
    \caption{Initial Image}
    \label{fig:cat}
\end{subfigure}
\hfill
\begin{subfigure}[b]{0.32\columnwidth}
    \includegraphics[width=\textwidth]{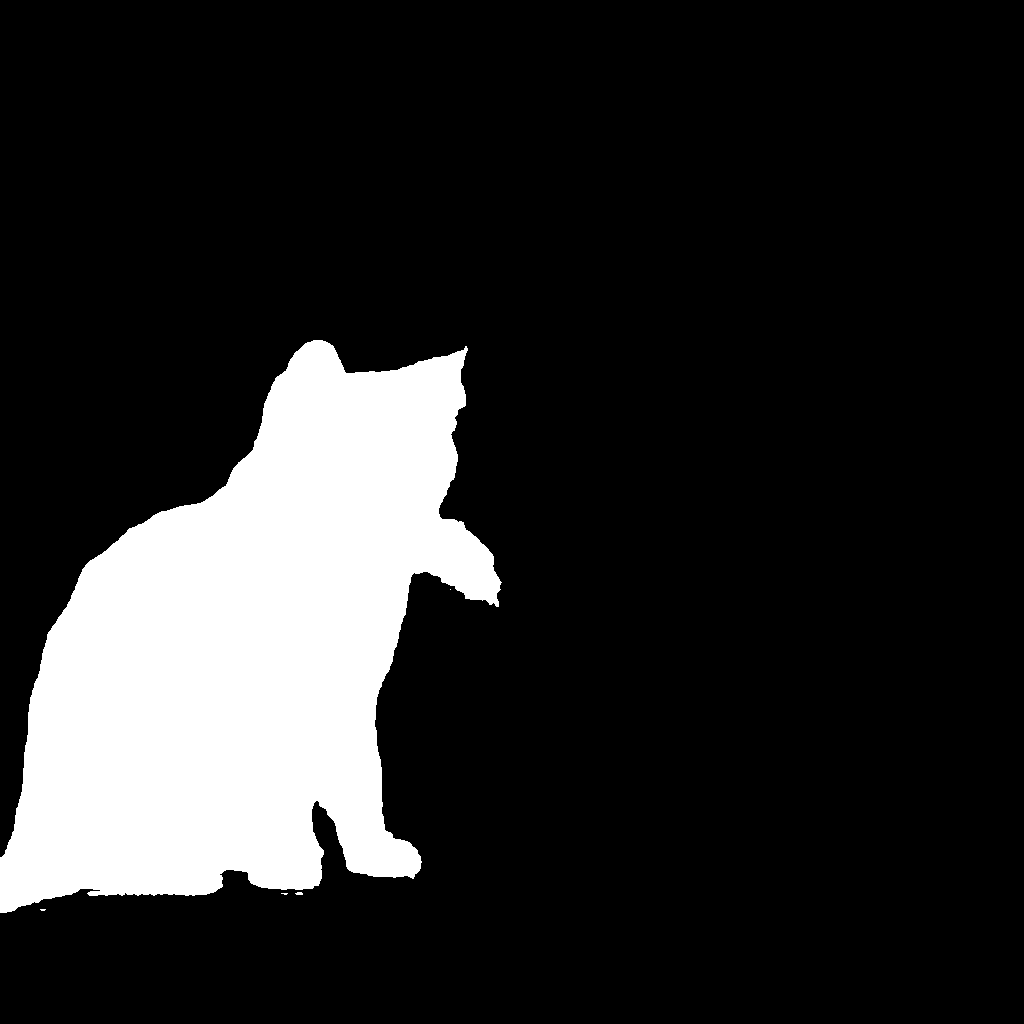}
    \caption{Mask Image}
    \label{fig:cat_mask}
\end{subfigure}
\hfill
\begin{subfigure}[b]{0.32\columnwidth}
    \includegraphics[width=\textwidth]{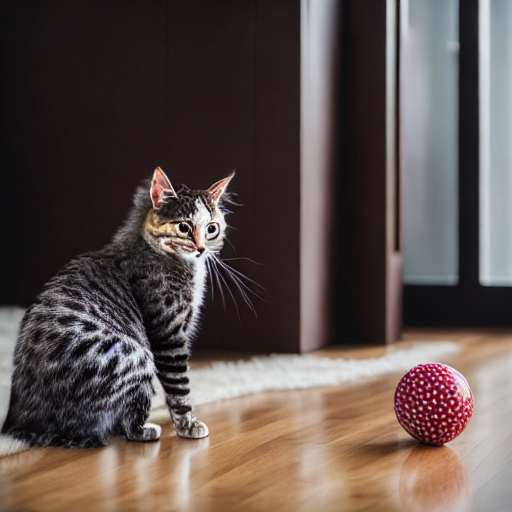}
    \caption{Inpainted Image}
    \label{fig:cat_inpainted}
\end{subfigure}
\caption{Example 3: a cat with a polka-dotted fur pattern}
\label{sample3}
\end{figure}

Figures \ref{sample2} and \ref{sample3} further demonstrate the effectiveness of our framework in rectifying biases. Figures \ref{sample2} present another instance of model bias, where the text-to-image model rendered diamonds as large and intact, contrary to the user's intent to depict broken diamonds. Similarly, Figure \ref{sample3} presents an initial image of a conventional cat, which did not match the user's unique request for a cat with a polka-dotted fur pattern. The framework effectively adjusted both images to align more closely with the user's specifications, introducing broken features to the diamonds and the distinct polka-dot pattern to the cat's fur.

We quantitatively validated our framework's effectiveness by comparing the CLIP scores \cite{radford2021learning}  of original and inpainted images, as shown in Table \ref{clip_score}. The increased CLIP scores for the inpainted images quantitatively confirm the framework's ability to enhance the correspondence between images and textual prompts, showcasing its effectiveness in refining image outputs to better reflect user intentions.

\begin{table*}[]
    \centering
    \begin{tabular}{c|c|c|c}
         \textbf{Figure No.} & \textbf{Prompt}  & \textbf{Initial Image}  & \textbf{Inpainted Image} \\
         \hline
         Figure \ref{sample1} & \small{A fantasy world where a river is made of dark chocolate}  & 20.441  & 20.518 \\
         \hline
         Figure \ref{sample2} & \small{Diamonds broke into pieces} & 22.653  & 22.824 \\
         \hline
         Figure \ref{sample3} & \small{A cat with a polka-dotted fur pattern} & 24.501  & 24.680 \\
    \end{tabular}
    \caption{CLIP Score Comparison}
    \label{clip_score}
\end{table*}

%% file: sec/6_conclusion.tex
\section{Conclusion}
This research aims to address the biases present in generative AI, which frequently produces images reflective of the predominant features in its training data, thereby struggling with less common or unconventional concepts. Our initial experiments showed that the accuracy with which text-to-image AI models depict uncommon concepts decreases when these concepts are combined with more familiar ones in a single prompt.

In response, we introduced the Inpaint Biases Framework, which employs inpainting techinique to selectively refine and correct the parts of an image that are not accurately represented. The process is directed by enhanced prompts from a Large Language Model (LLM), focusing on those parts of the image that are inaccurately represented.

Consequently, the framework not only improves the visual accuracy of AI-generated images but also addresses a critical ethical dimension by reducing the bias towards dominant data features. This study offers a technological pathway to diminish the biases in generative AI, demonstrating a commitment to ethical principles in AI development and application.

%% file: sec/7_future_research.tex
\section{Future Research}
\label{sec:future}
In this paper, we examined a specific bias inherent in text-to-image models and proposed the Inpaint Biases Framework as a preliminary solution to this issue. While the framework marks an important first step toward addressing biases in image generation, further research is necessary to refine and enhance its effectiveness. Future studies could explore methods whereby the model autonomously identifies and corrects misalignments between prompts and generated images without user intervention. For instance, the model could automatically detect areas with low alignment to the prompt or select the specific aspects of a prompt to refine. Such advancements would not only streamline the process but also potentially enhance the accuracy and relevance of the corrections. We hope that future research will continue to advance text-to-image models in a direction that systematically addresses and mitigates biases, further empowering these models as unbiased tools for creative expression.

%% file: main.bbl
\begin{thebibliography}{6}
\providecommand{\natexlab}[1]{#1}
\providecommand{\url}[1]{\texttt{#1}}
\expandafter\ifx\csname urlstyle\endcsname\relax
  \providecommand{\doi}[1]{doi: #1}\else
  \providecommand{\doi}{doi: \begingroup \urlstyle{rm}\Url}\fi

\bibitem[Kirillov et~al.(2023)Kirillov, Mintun, Ravi, Mao, Rolland, Gustafson, Xiao, Whitehead, Berg, Lo, Dollár, and Girshick]{kirillov2023segment}
Alexander Kirillov, Eric Mintun, Nikhila Ravi, Hanzi Mao, Chloe Rolland, Laura Gustafson, Tete Xiao, Spencer Whitehead, Alexander~C. Berg, Wan-Yen Lo, Piotr Dollár, and Ross Girshick.
\newblock Segment anything, 2023.

\bibitem[Radford et~al.(2021)Radford, Kim, Hallacy, Ramesh, Goh, Agarwal, Sastry, Askell, Mishkin, Clark, Krueger, and Sutskever]{radford2021learning}
Alec Radford, Jong~Wook Kim, Chris Hallacy, Aditya Ramesh, Gabriel Goh, Sandhini Agarwal, Girish Sastry, Amanda Askell, Pamela Mishkin, Jack Clark, Gretchen Krueger, and Ilya Sutskever.
\newblock Learning transferable visual models from natural language supervision, 2021.

\bibitem[Ramesh et~al.(2021)Ramesh, Pavlov, Goh, Gray, Voss, Radford, Chen, and Sutskever]{ramesh2021zeroshot}
Aditya Ramesh, Mikhail Pavlov, Gabriel Goh, Scott Gray, Chelsea Voss, Alec Radford, Mark Chen, and Ilya Sutskever.
\newblock Zero-shot text-to-image generation, 2021.

\bibitem[Rombach et~al.(2022)Rombach, Blattmann, Lorenz, Esser, and Ommer]{rombach2022highresolution}
Robin Rombach, Andreas Blattmann, Dominik Lorenz, Patrick Esser, and Björn Ommer.
\newblock High-resolution image synthesis with latent diffusion models, 2022.

\bibitem[Saharia et~al.(2022)Saharia, Chan, Saxena, Li, Whang, Denton, Ghasemipour, Ayan, Mahdavi, Lopes, Salimans, Ho, Fleet, and Norouzi]{saharia2022photorealistic}
Chitwan Saharia, William Chan, Saurabh Saxena, Lala Li, Jay Whang, Emily Denton, Seyed Kamyar~Seyed Ghasemipour, Burcu~Karagol Ayan, S.~Sara Mahdavi, Rapha~Gontijo Lopes, Tim Salimans, Jonathan Ho, David~J Fleet, and Mohammad Norouzi.
\newblock Photorealistic text-to-image diffusion models with deep language understanding, 2022.

\bibitem[Suvorov et~al.(2021)Suvorov, Logacheva, Mashikhin, Remizova, Ashukha, Silvestrov, Kong, Goka, Park, and Lempitsky]{suvorov2021resolutionrobust}
Roman Suvorov, Elizaveta Logacheva, Anton Mashikhin, Anastasia Remizova, Arsenii Ashukha, Aleksei Silvestrov, Naejin Kong, Harshith Goka, Kiwoong Park, and Victor Lempitsky.
\newblock Resolution-robust large mask inpainting with fourier convolutions, 2021.

\end{thebibliography}
